\pdfoutput=1
\documentclass{article} 
\usepackage{times}

\usepackage[letterpaper,margin=1in]{geometry}
\usepackage{titling}
\usepackage[round]{natbib}

\usepackage{amsmath,amsfonts,bm}

\def\eqref#1{equation~\ref{#1}}

\def\1{\bm{1}}

\DeclareMathAlphabet{\mathsfit}{\encodingdefault}{\sfdefault}{m}{sl}
\SetMathAlphabet{\mathsfit}{bold}{\encodingdefault}{\sfdefault}{bx}{n}

\usepackage{hyperref}
\usepackage{url}
\usepackage{booktabs}
\usepackage{amsfonts}
\usepackage{amsmath}
\usepackage{graphicx}
\usepackage{subcaption}
\usepackage{multirow}
\usepackage{array}
\usepackage{xcolor}

\usepackage{array}

\title{Benchmarking Open-Source Safety Guard Models: \\ A Comprehensive Evaluation\thanks{Published as a workshop paper at ICLR 2026}}

\author{%
\begin{tabular}{>{\centering\arraybackslash}p{0.32\textwidth} >{\centering\arraybackslash}p{0.32\textwidth} >{\centering\arraybackslash}p{0.32\textwidth}}
\textbf{Reetu Raj Harsh} & \textbf{Bhaskarjit Sarmah} & \textbf{Stefano Pasquali} \\
Domyn & Domyn & Domyn \\
Gurugram, India & Gurugram, India & New York, USA \\
{\small\url{reeturaj.harsh@domyn.com}} & {\small\url{bhaskarjit.sarmah@domyn.com}} & {\small\url{stefano.pasquali@domyn.com}}
\end{tabular}
}

\date{}

\begin{document}

\maketitle

\begin{abstract}
As Large Language Models (LLMs) are increasingly deployed in safety-critical applications, robust content moderation becomes essential. We present a comprehensive evaluation of 14 open-source safety guard models on a curated benchmark of 79,331 samples spanning 8 NIST AI Risk Framework safety categories. Our benchmark aggregates four diverse datasets (HarmBench, StrongREJECT, RealToxicityPrompts, and BeaverTails), filtered to focus exclusively on safety-relevant content (violence, hate speech, harassment, sexual content, suicide/self-harm, profanity, threats, and health misinformation). We find that \textbf{recall is the critical metric} for safety applications, as missing harmful content poses greater risk than false positives. Our evaluation reveals surprising results: Qwen Guard (4B parameters) achieves the highest recall (83.97\%) while larger models like Llama Guard (12B) and GPT-OSS Safeguard (20B) exhibit conservative behavior, missing up to 75\% of unsafe content. We demonstrate that model size does not correlate with safety detection performance and that general-purpose guard models outperform specialized ones. These findings provide practical guidance for selecting safety guard models in production deployments.
\end{abstract}

\section{Introduction}
\label{sec:introduction}

The rapid adoption of Large Language Models (LLMs) in various applications introduces significant safety concerns that demand rigorous content moderation. Safety guard models have emerged as a critical component in LLM deployment pipelines, serving as filters that classify input prompts or model outputs as safe or unsafe. However, the proliferation of guard models, each with different architectures, training methodologies, and safety taxonomies, creates uncertainty for practitioners selecting appropriate safeguards for their applications.

Despite the critical importance of guard models, no comprehensive benchmark exists that evaluates modern open-source safety models across diverse, safety-focused datasets using standardized taxonomy.

\subsection{Research Questions}

This work addresses three primary research questions:

\begin{enumerate}
    \item \textbf{RQ1}: How do existing open-source guard models perform on safety detection across diverse datasets and harm categories?
    \item \textbf{RQ2}: What is the relationship between model size, architecture, and safety detection performance?
    \item \textbf{RQ3}: Which evaluation metrics best capture guard model effectiveness for safety-critical deployments?
\end{enumerate}

\subsection{Contributions}

Our contributions include: (1) a comprehensive benchmark of 79,331 samples from four publicly available sources, filtered using the NIST AI Risk Management Framework \citep{nist2023airmf} to focus on 8 safety subcategories; (2) large-scale evaluation of 14 open-source guard models (110M--20B parameters) from Google, NVIDIA, IBM, Meta, and Alibaba; (3) metric analysis demonstrating that recall is the primary metric and high-precision models can be dangerously conservative; and (4) actionable recommendations for model selection.

\section{Related Work}
\label{sec:related_work}

\subsection{Guard Models}

Safety guard models have evolved from simple keyword-based filters to sophisticated LLM-based classifiers. Llama Guard \citep{meta2025llamaguard4} introduced a taxonomy-driven approach using instruction-tuned models for input-output safeguarding. WildGuard \citep{wildguard2024} extended this to handle both prompt and response classification with broader coverage. More recently, models from diverse organizations have emerged: Granite Guardian \citep{padhi2024graniteguardian} from IBM, Qwen Guard \citep{zhao2025qwen3guard} from Alibaba, and ShieldGemma \citep{google2024shieldgemma} from Google, each with varying safety taxonomies and architectural choices.

\subsection{Safety Benchmarks}

As large language models become more powerful and widely deployed, ensuring the safety of their outputs is critical. To mitigate risks, \textit{guardrail models} such as Llama Guard, ShieldGemma, WildGuard, and Qwen Guard are adopted as filtering mechanisms that perform real-time risk detection on both user prompts and model responses, ensuring safer interactions in AI systems.

Existing safety benchmarks primarily target general-purpose LLMs. SafetyBench \citep{zhang2024safetybench} provides 11,435 multiple-choice questions to test whether LLMs select safe responses. HELM \citep{liang2023helm} evaluates LLMs holistically across many dimensions including safety. RabakBench \citep{chua2025rabakbench} constructs localized multilingual safety benchmarks for low-resource languages. These benchmarks evaluate the generation behavior of LLMs, not the effectiveness of guardrail models.

For guardrail model evaluation, GuardBench \citep{bassani2024guardbench} introduced a benchmark comprising 40 datasets and evaluated 13 models. However, their evaluated models were primarily Llama Guard variants and content moderation models (Detoxify, ToxiGen), and they acknowledge the lack of ``a generally accepted taxonomy of unsafe content,'' leaving results fragmented across datasets without unified categorization.

Our work focuses on evaluating guardrail models and addresses these gaps: (1) we evaluate 14 models from diverse vendors (Google, NVIDIA, IBM, Meta, Alibaba) including recent 2025 models (Qwen Guard, GPT-OSS Safeguard, DynaGuard, GuardReasoner, Llama Guard 4); (2) we adopt the NIST AI Risk Management Framework \citep{nist2023airmf} for standardized categorization across 8 safety subcategories; and (3) we provide category-level analysis revealing which safety categories remain challenging for current guardrail models.

\section{Methodology}
\label{sec:methodology}

\subsection{Dataset Construction}
\label{sec:dataset}

We construct a master safety dataset by aggregating and filtering four publicly available sources according to the NIST AI Risk Management Framework.

\paragraph{NIST AI Risk Framework.}
The NIST AI RMF defines 7 risk categories: SAFETY, SECURITY, ACCOUNTABILITY, PRIVACY, FAIRNESS, TRANSPARENCY, and UNCERTAINTY. Our benchmark focuses exclusively on SAFETY to ensure evaluation targets safety-specific detection rather than conflating safety with security or privacy. Within SAFETY, we define 8 subcategories: (1) Violence, (2) Hate Speech, (3) Suicide \& Self-Harm, (4) Sexual Content, (5) Harassment, (6) Profanity, (7) Threats, and (8) Health misinformation.

\paragraph{Source Datasets.}

\textbf{HarmBench} \citep{mazeika2024harmbench}: 400 text behaviors designed to test model robustness. We retain 103 samples after filtering out copyright (100), cybercrime and illegal activity (132 $\rightarrow$ SECURITY), and misinformation (65 $\rightarrow$ ACCOUNTABILITY).

\textbf{StrongREJECT} \citep{souly2024strongreject}: 313 forbidden prompts testing rejection capabilities. We retain 154 samples (Violence, Sexual Content, Hate/Harassment) after filtering out non-violent crimes and illegal goods (109 $\rightarrow$ SECURITY) and disinformation (50 $\rightarrow$ ACCOUNTABILITY).

\textbf{RealToxicityPrompts} \citep{gehman2020realtoxicityprompts}: 99,442 naturally occurring prompts with Perspective API toxicity scores across 8 dimensions. We retain 67,521 samples after filtering out the \textit{flirtation} category (31,921 samples) as it does not map to any of our 8 NIST SAFETY subcategories. We map remaining scores to binary labels using a threshold of 0.5, the standard midpoint for binary classification that balances false positives and false negatives.

\textbf{BeaverTails} \citep{beavertails}: 27,186 human-annotated samples with 14 harm categories. We retain 11,553 samples after filtering out terrorism and weapons (SECURITY), financial crime and privacy violations (PRIVACY), political content (CROSS-CUTTING), and non-violent unethical behavior (not safety-specific).

\paragraph{Dataset Selection Rationale.}
Our dataset achieves complete coverage of all 8 NIST safety subcategories: RealToxicityPrompts provides naturally-occurring toxicity (Harassment, Threats, Profanity, Hate Speech); BeaverTails contributes human-annotated samples (Violence, Sexual Content, Health, and critically, Suicide \& Self-Harm-the only source for this category); HarmBench and StrongREJECT add adversarial edge cases. This 79,331-sample benchmark spans adversarial, natural, and human-annotated sources.

\paragraph{Labeling Logic.}
Labeling varies by dataset: HarmBench and StrongREJECT samples are all \textit{unsafe} (adversarial by design); BeaverTails uses human-annotated labels; RealToxicityPrompts computes the maximum toxicity score across 7 dimensions (excluding flirtation) and labels samples as unsafe if score $> 0.5$. Full labeling details are in Appendix~\ref{app:implementation}.

\begin{table}[t]
\caption{Master Safety Dataset Composition}
\label{tab:dataset_composition}
\begin{center}
\begin{tabular}{lccccc}
\toprule
\textbf{Dataset} & \textbf{Original} & \textbf{Filtered} & \textbf{Safe} & \textbf{Unsafe} & \textbf{Labeling} \\
\midrule
HarmBench & 400 & 103 & 0 & 103 & All unsafe \\
StrongREJECT & 313 & 154 & 0 & 154 & All unsafe \\
RealToxicityPrompts & 99,442 & 67,521 & 35,938 & 31,583 & Threshold 0.5 \\
BeaverTails & 27,186 & 11,553 & 0 & 11,553 & Human-annotated \\
\midrule
\textbf{Total} & \textbf{127,341} & \textbf{79,331} & \textbf{35,938} & \textbf{43,393} & - \\
\bottomrule
\end{tabular}
\end{center}
\end{table}

The final dataset contains 79,331 samples with 54.7\% unsafe and 45.3\% safe labels. Notably, all safe samples originate from RealToxicityPrompts, as the other three datasets are designed exclusively for adversarial/harmful content evaluation.

\subsection{Guard Models Evaluated}
\label{sec:models}

We evaluate 14 open-source guard models spanning diverse architectures, sizes, and safety taxonomies. Table~\ref{tab:models} summarizes the models.

\begin{table}[t]
\caption{Guard Models Evaluated. Models are grouped by architecture type: decoder-only LLMs (top) and encoder-only transformers (bottom).}
\label{tab:models}
\begin{center}
\small
\begin{tabular}{lclc}
\toprule
\textbf{Model} & \textbf{Size} & \textbf{Base Architecture} & \textbf{Categories} \\
\midrule
\multicolumn{4}{l}{\textit{Decoder-only LLMs}} \\
\midrule
Qwen Guard \citep{zhao2025qwen3guard} & 4B & Qwen3 & 10 \\
Nemotron Safety \citep{joshi2025cultureguard} & 8B & Llama 3.1 & 23 \\
WildGuard \citep{wildguard2024} & 7B & Mistral-7B & 13 \\
MD-Judge \citep{li2024salad} & 7B & InternLM2 & 16 \\
Granite Guardian \citep{padhi2024graniteguardian} & 8B & Granite & Custom \\
DynaGuard \citep{hoover2025dynaguard} & 8B & Qwen3 & Dynamic \\
DuoGuard \citep{deng2025duoguard} & 0.5B & Qwen 2.5 & 12 \\
Llama Guard \citep{meta2025llamaguard4} & 12B & Llama 4 (pruned) & 14 \\
ShieldGemma \citep{google2024shieldgemma} & 2B & Gemma 2 & 4 \\
GuardReasoner \citep{liu2025guardreasoner} & 3B & Llama 3.2 & Reasoning \\
GPT-OSS Safeguard \citep{openai2025gptoss} & 20B & GPT-OSS & Custom \\
\midrule
\multicolumn{4}{l}{\textit{Encoder-only Transformers}} \\
\midrule
EthicalEye \citep{patel2024ethicaleye} & 270M & XLM-RoBERTa & Binary \\
PoliteGuard \citep{intel2024politeguard} & 110M & BERT-base & 4 \\
MetaHateBERT \citep{piot2024metahate} & 110M & BERT-base & Binary \\
\bottomrule
\end{tabular}
\end{center}
\end{table}

\paragraph{Label Normalization.} Models output varying labels (safe, unsafe, controversial, error). We normalize by mapping Qwen Guard's \textit{controversial} label to \textit{unsafe} (as these contain contextually harmful content that should be flagged), excluding error predictions, and mapping all other labels to binary safe/unsafe.

\section{Evaluation and Results}
\label{sec:results}

\subsection{Evaluation Metrics (RQ3)}

For safety-critical applications, recall is the primary metric: missing harmful content (false negatives) poses greater risk than incorrectly flagging safe content (false positives). We report Recall, Precision, F1 Score, Accuracy, ROC-AUC, and MCC (Matthews Correlation Coefficient), with results ranked by recall throughout.

\subsection{Overall Performance (RQ1)}

Table~\ref{tab:main_results} presents the main evaluation results, ranked by recall. Our primary finding is that Qwen Guard achieves the highest recall (83.97\%), significantly outperforming larger models.

\begin{table}[t]
\caption{Guard Model Performance (Ranked by Recall)}
\label{tab:main_results}
\begin{center}
\small
\begin{tabular}{clccccc}
\toprule
\textbf{Rank} & \textbf{Model} & \textbf{Size} & \textbf{Recall} & \textbf{Precision} & \textbf{F1} & \textbf{Accuracy} \\
\midrule
1 & Qwen Guard & 4B & \textbf{0.8397} & 0.6879 & 0.7563 & 0.7036 \\
2 & Nemotron Safety & 8B & 0.7725 & 0.7493 & 0.7607 & 0.7342 \\
3 & WildGuard & 7B & 0.7383 & 0.7289 & 0.7336 & 0.7067 \\
4 & MD-Judge & 7B & 0.7091 & 0.7316 & 0.7202 & 0.6986 \\
5 & Granite Guardian & 8B & 0.6881 & 0.7677 & 0.7257 & 0.7155 \\
6 & DynaGuard & 8B & 0.6659 & 0.7715 & 0.7148 & 0.7095 \\
7 & DuoGuard & 0.5B & 0.6360 & 0.7612 & 0.6930 & 0.6918 \\
8 & EthicalEye & 270M & 0.6032 & 0.6705 & 0.6350 & 0.6208 \\
9 & PoliteGuard & 110M & 0.5873 & 0.5689 & 0.5780 & 0.5308 \\
10 & GuardReasoner & 3B & 0.5109 & 0.7397 & 0.6044 & 0.6283 \\
11 & ShieldGemma & 2B & 0.4549 & 0.8220 & 0.5857 & 0.6479 \\
12 & Llama Guard & 12B & 0.3332 & 0.7851 & 0.4679 & 0.5860 \\
13 & GPT-OSS Safeguard & 20B & 0.2486 & 0.8068 & 0.3800 & 0.5564 \\
14 & MetaHateBERT & 110M & 0.1579 & 0.6541 & 0.2544 & 0.4937 \\
\bottomrule
\end{tabular}
\end{center}
\end{table}

Key observations: (1) Conservative models are dangerous-while ShieldGemma achieves highest precision (82.20\%), it misses 54.51\% of unsafe content, and GPT-OSS Safeguard misses 75.14\%; (2) General-purpose models outperform specialized ones-MetaHateBERT, designed for hate speech, achieves only 15.79\% recall, failing to generalize across categories.

\begin{figure}[t]
\centering
\begin{subfigure}[b]{0.48\textwidth}
    \centering
    \includegraphics[width=\textwidth]{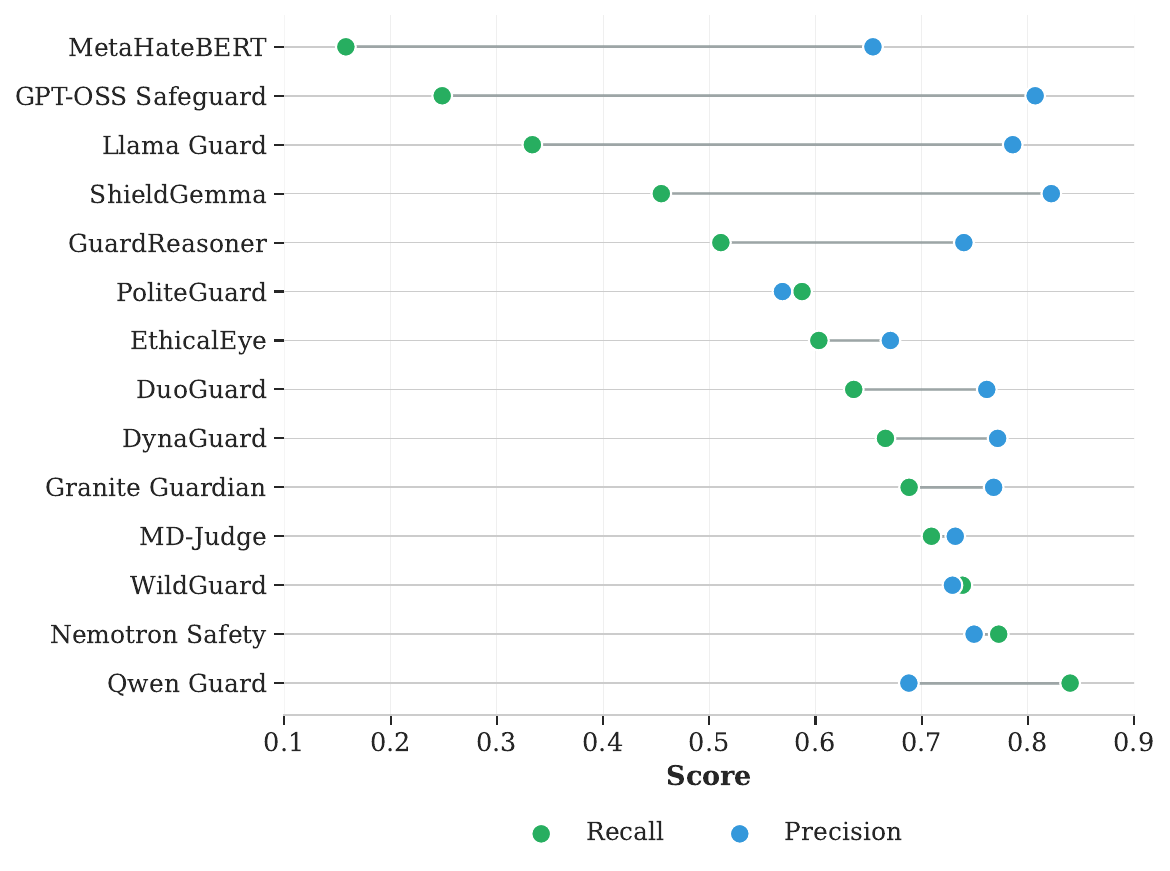}
    \caption{Precision-Recall Tradeoff}
    \label{fig:precision_recall}
\end{subfigure}
\hfill
\begin{subfigure}[b]{0.48\textwidth}
    \centering
    \includegraphics[width=\textwidth]{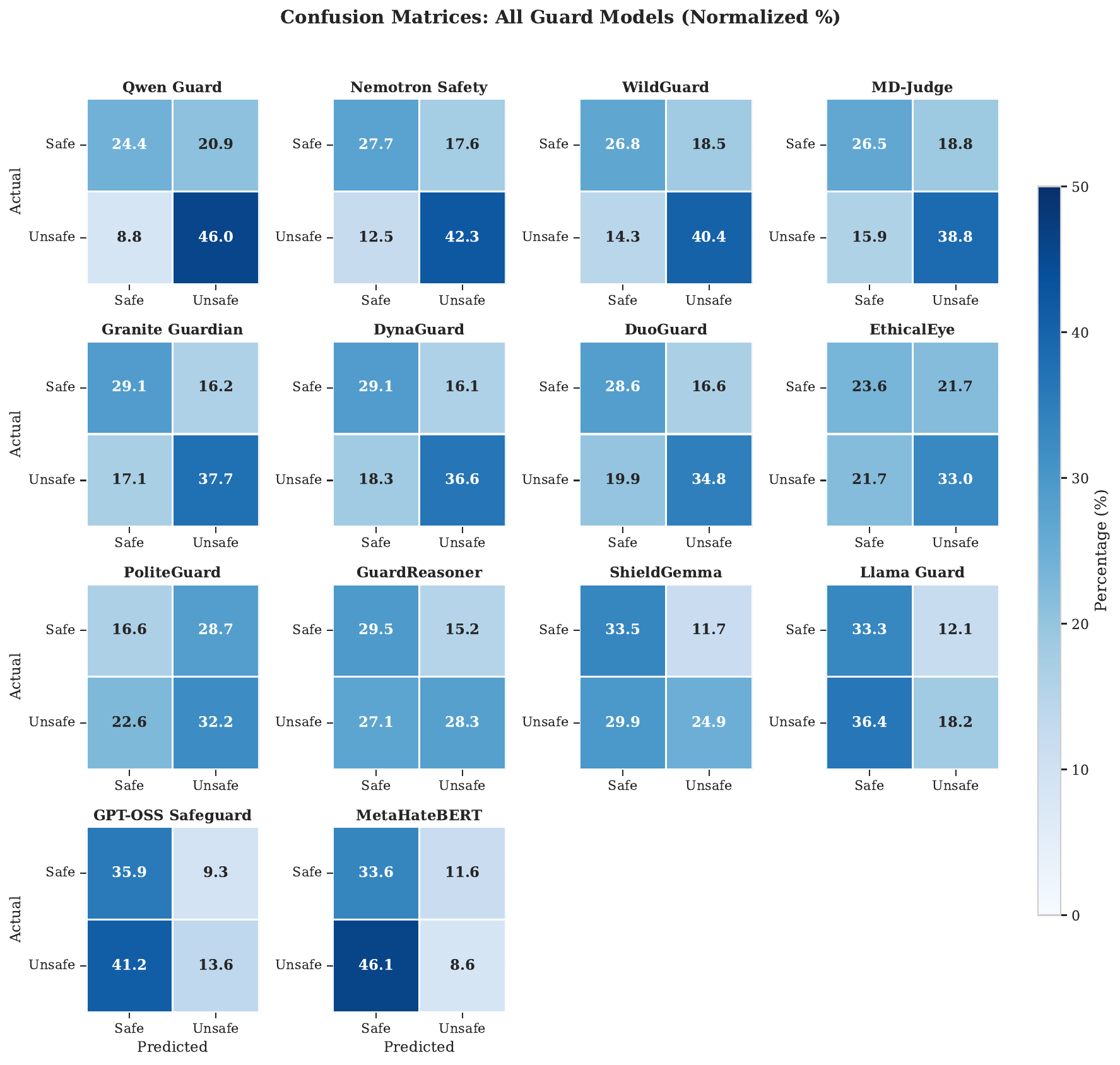}
    \caption{Confusion Matrices}
    \label{fig:confusion_matrices}
\end{subfigure}
\caption{Performance analysis of 14 guard models. \textbf{(a)} Precision-recall comparison where each line connects recall (green) to precision (blue). Line length indicates the gap-models with long leftward lines (GPT-OSS, Llama Guard) miss most unsafe content despite high precision. \textbf{(b)} Normalized confusion matrices sorted by recall. The bottom-left cell (FN) shows missed unsafe content: top models achieve 8.8--12.5\% FN, while conservative models show 41--46\% FN rates.}
\label{fig:main_results}
\end{figure}

Figure~\ref{fig:main_results} visualizes model performance through two complementary views: the precision-recall tradeoff and detailed confusion matrices.

\subsection{Performance by Dataset}

Table~\ref{tab:dataset_performance} shows per-dataset recall for the top 5 models.

\begin{table}[t]
\caption{Recall by Dataset (Top 5 Models)}
\label{tab:dataset_performance}
\begin{center}
\small
\begin{tabular}{lcccc}
\toprule
\textbf{Model} & \textbf{HarmBench} & \textbf{StrongREJECT} & \textbf{RealToxicity} & \textbf{BeaverTails} \\
\midrule
Qwen Guard & 1.0000 & 0.5455 & 0.8004 & 0.9477 \\
Nemotron Safety & 0.9903 & 1.0000 & 0.7303 & 0.8828 \\
WildGuard & 1.0000 & 1.0000 & 0.6656 & 0.9313 \\
MD-Judge & 0.9903 & 1.0000 & 0.6215 & 0.9424 \\
Granite Guardian & 0.9903 & 1.0000 & 0.6243 & 0.8559 \\
\bottomrule
\end{tabular}
\end{center}
\end{table}

Adversarial datasets show varied performance: Most models achieve near-perfect recall on HarmBench (99-100\%), while StrongREJECT performance varies-Qwen Guard achieves only 54.55\% recall on StrongREJECT despite leading overall, suggesting dataset-specific biases. RealToxicityPrompts proves most challenging across all models, containing subtle, naturally-occurring toxicity that many models fail to detect. To verify that results are not driven by source artifacts, we provide stratified analysis comparing RealToxicityPrompts (balanced safe/unsafe from same source) versus adversarial datasets combined in Appendix~\ref{app:per_source}; model rankings remain consistent across both splits.

\subsection{Performance by NIST Category}

Table~\ref{tab:category_recall} shows recall variation across safety categories for all models. This analysis reveals systematic patterns in what types of harmful content models detect well versus poorly.

\begin{table}[t]
\caption{Recall by NIST Safety Category (All 14 Models). Categories ordered by detection difficulty (avg recall): Suicide/Self-Harm (78\%) $>$ Violence (71\%) $>$ Hate (62\%) $>$ Sexual (59\%) $>$ Health (57\%) $>$ Harassment (54\%) $>$ Profanity (51\%) $>$ Threats (43\%). Bold indicates best per category.}
\label{tab:category_recall}
\begin{center}
\scriptsize
\begin{tabular}{lcccccccc}
\toprule
\textbf{Model} & \textbf{Viol.} & \textbf{Suicide} & \textbf{Hate} & \textbf{Sexual} & \textbf{Health} & \textbf{Harass.} & \textbf{Prof.} & \textbf{Threat} \\
\midrule
Qwen Guard & 0.980 & 0.977 & \textbf{0.892} & \textbf{0.891} & 0.822 & 0.778 & 0.763 & \textbf{0.726} \\
Nemotron Safety & 0.956 & 0.931 & 0.708 & 0.830 & 0.738 & 0.759 & 0.850 & 0.612 \\
WildGuard & 0.969 & 0.960 & 0.811 & 0.729 & 0.778 & 0.704 & 0.560 & 0.577 \\
MD-Judge & \textbf{0.989} & \textbf{0.994} & 0.807 & 0.665 & \textbf{0.856} & 0.572 & 0.431 & 0.676 \\
Granite Guardian & 0.901 & 0.931 & 0.756 & 0.770 & 0.684 & 0.538 & 0.830 & 0.516 \\
DynaGuard & 0.913 & 0.948 & 0.760 & 0.683 & 0.731 & 0.538 & 0.480 & 0.565 \\
DuoGuard & 0.841 & 0.884 & 0.642 & 0.639 & 0.679 & 0.761 & 0.766 & 0.238 \\
GuardReasoner & 0.877 & 0.923 & 0.608 & 0.500 & 0.723 & 0.287 & 0.189 & 0.507 \\
EthicalEye & 0.186 & 0.416 & 0.483 & 0.775 & 0.215 & \textbf{0.895} & \textbf{0.923} & 0.442 \\
PoliteGuard & 0.401 & 0.532 & 0.696 & 0.403 & 0.344 & 0.840 & 0.826 & 0.442 \\
ShieldGemma & 0.697 & 0.960 & 0.650 & 0.643 & 0.548 & 0.165 & 0.144 & 0.343 \\
Llama Guard & 0.807 & 0.815 & 0.328 & 0.297 & 0.574 & 0.186 & 0.121 & 0.205 \\
GPT-OSS Safeguard & 0.365 & 0.613 & 0.341 & 0.268 & 0.220 & 0.229 & 0.086 & 0.114 \\
MetaHateBERT & 0.036 & 0.046 & 0.202 & 0.112 & 0.056 & 0.290 & 0.206 & 0.085 \\
\bottomrule
\end{tabular}
\end{center}
\end{table}

Category-specific findings reveal systematic patterns: Violence and Suicide/Self-Harm are easiest to detect ($>$90\% recall for most LLMs), while Threats is hardest (43.2\% average). Encoder models show surprising strength-EthicalEye achieves highest recall on Harassment (89.5\%) and Profanity (92.3\%) despite ranking 8th overall. No single model dominates: MD-Judge leads on Violence/Suicide/Health, Qwen Guard on Hate/Sexual/Threats. Large models (GPT-OSS, Llama Guard) consistently underperform on implicit harm categories.

\subsection{Threshold Sensitivity Analysis}

Using the labeling function defined in Section~\ref{sec:dataset}, we analyze sensitivity to threshold $\tau \in \{0.3, 0.4, 0.5, 0.6, 0.7\}$ for RealToxicityPrompts. Table~\ref{tab:threshold_main} shows results for top 5 models.

\begin{table}[t]
\caption{Threshold Sensitivity: Recall (R), Precision (P), and F1 at Different Labeling Thresholds. Bold indicates default threshold 0.5. At $\tau$=0.3, 90.6\% of samples are unsafe; at $\tau$=0.7, only 24.5\% are unsafe.}
\label{tab:threshold_main}
\begin{center}
\scriptsize
\begin{tabular}{llccccc}
\toprule
\textbf{Model} & \textbf{Metric} & \textbf{0.3} & \textbf{0.4} & \textbf{0.5} & \textbf{0.6} & \textbf{0.7} \\
\midrule
\multirow{3}{*}{Qwen Guard}
    & R & 0.712 & 0.757 & \textbf{0.840} & 0.898 & 0.923 \\
    & P & 0.965 & 0.899 & \textbf{0.688} & 0.431 & 0.338 \\
    & F1 & 0.819 & 0.822 & \textbf{0.756} & 0.582 & 0.494 \\
\midrule
\multirow{3}{*}{Nemotron Safety}
    & R & 0.613 & 0.675 & \textbf{0.773} & 0.836 & 0.868 \\
    & P & 0.985 & 0.949 & \textbf{0.749} & 0.474 & 0.376 \\
    & F1 & 0.756 & 0.789 & \textbf{0.761} & 0.605 & 0.525 \\
\midrule
\multirow{3}{*}{WildGuard}
    & R & 0.596 & 0.644 & \textbf{0.738} & 0.827 & 0.868 \\
    & P & 0.974 & 0.921 & \textbf{0.729} & 0.478 & 0.383 \\
    & F1 & 0.739 & 0.758 & \textbf{0.734} & 0.606 & 0.531 \\
\midrule
\multirow{3}{*}{MD-Judge}
    & R & 0.567 & 0.609 & \textbf{0.709} & 0.816 & 0.860 \\
    & P & 0.968 & 0.910 & \textbf{0.732} & 0.493 & 0.396 \\
    & F1 & 0.715 & 0.729 & \textbf{0.720} & 0.614 & 0.543 \\
\midrule
\multirow{3}{*}{Granite Guardian}
    & R & 0.533 & 0.588 & \textbf{0.688} & 0.779 & 0.832 \\
    & P & 0.985 & 0.952 & \textbf{0.768} & 0.509 & 0.414 \\
    & F1 & 0.692 & 0.727 & \textbf{0.726} & 0.616 & 0.553 \\
\bottomrule
\end{tabular}
\end{center}
\end{table}

Key findings: (1) Model rankings are stable-Qwen Guard maintains highest recall at all thresholds, followed by Nemotron Safety and WildGuard. (2) Precision-recall tradeoff-as threshold increases (stricter ground truth), recall improves while precision degrades. (3) F1 peaks at 0.5-justifying our default threshold choice. Full results for all 14 models in Appendix~\ref{app:threshold_sensitivity}.

\subsection{Error Analysis: False Negatives}
\label{sec:error_analysis}

To understand model failures beyond aggregate metrics, we analyze false negatives-unsafe content incorrectly classified as safe. This analysis reveals systematic blind spots that explain the 5.3$\times$ recall gap between best and worst performers.

\textbf{Methodology.} For each of the 43,393 unsafe samples, we compute each model's prediction and calculate false negative rates (FN rate = samples missed / total unsafe) by NIST category. Table~\ref{tab:fn_analysis} shows results for top-3 and bottom-3 models by overall recall.

\begin{table}[t]
\caption{False Negative Rate (\%) by NIST Category. Lower is better. Top-3 models (Qwen, Nemotron, WildGuard) vs Bottom-3 (Llama Guard, GPT-OSS, MetaHateBERT). Bold indicates best per category. Full 14-model results in Appendix~\ref{app:fn_analysis}.}
\label{tab:fn_analysis}
\begin{center}
\scriptsize
\begin{tabular}{l|ccc|ccc}
\toprule
& \multicolumn{3}{c|}{\textbf{Top 3 Models}} & \multicolumn{3}{c}{\textbf{Bottom 3 Models}} \\
\textbf{Category} & \textbf{Qwen} & \textbf{Nemotron} & \textbf{WildGuard} & \textbf{Llama} & \textbf{GPT-OSS} & \textbf{MetaHateBERT} \\
\midrule
Violence & \textbf{1.9} & 4.4 & 3.1 & 19.3 & 63.5 & 96.4 \\
Suicide/Self-Harm & 2.3 & 6.9 & 4.0 & 18.5 & 38.7 & 95.4 \\
Hate Speech & \textbf{10.3} & 29.3 & 18.9 & 67.2 & 65.9 & 79.8 \\
Sexual Content & \textbf{10.6} & 17.0 & 27.1 & 70.3 & 73.2 & 88.8 \\
Health & 17.8 & 26.2 & 22.2 & 42.6 & 78.0 & 94.4 \\
Harassment & 22.2 & 24.2 & 29.6 & 81.4 & 77.1 & 71.0 \\
Profanity & 23.8 & \textbf{15.1} & 44.1 & 87.9 & 91.4 & 79.4 \\
Threats & \textbf{27.4} & 38.8 & 42.3 & 79.5 & 88.6 & 91.5 \\
\midrule
\textbf{Overall} & \textbf{15.9} & 22.8 & 26.2 & 66.7 & 75.2 & 84.2 \\
\bottomrule
\end{tabular}
\end{center}
\end{table}

Key findings: The 5.3$\times$ performance gap between Qwen Guard (15.9\% FN) and MetaHateBERT (84.2\% FN) reveals systematic differences. Consistent with Section 4.3, Suicide/Self-Harm has lowest FN rate (21.8\%) while Threats has highest (56.1\%). Conservative models show a striking explicit-vs-implicit pattern: Llama Guard achieves 18.9\% FN on explicit categories but 83.0\% on implicit categories (64.1pp gap), suggesting reliance on keyword matching rather than semantic understanding.

\section{Discussion}
\label{sec:discussion}

Our evaluation addresses the three research questions posed in Section~\ref{sec:introduction}: RQ1 is answered through our comprehensive performance analysis across datasets and NIST categories (Sections 4.2--4.5), RQ2 through our model size analysis (Finding 2), and RQ3 through our metric analysis demonstrating recall as the critical metric (Section 4.1, Finding 1).

\subsection{Key Findings}

\textbf{Finding 1 (RQ3): High precision does not compensate for low recall.}

Precision-optimized models (GPT-OSS Safeguard, Llama Guard) miss up to 75\% of unsafe content. A 25\% recall rate is insufficient regardless of precision.

\textbf{Finding 2 (RQ2): Larger is not safer.}

\begin{figure}[t]
\centering
\includegraphics[width=0.85\columnwidth]{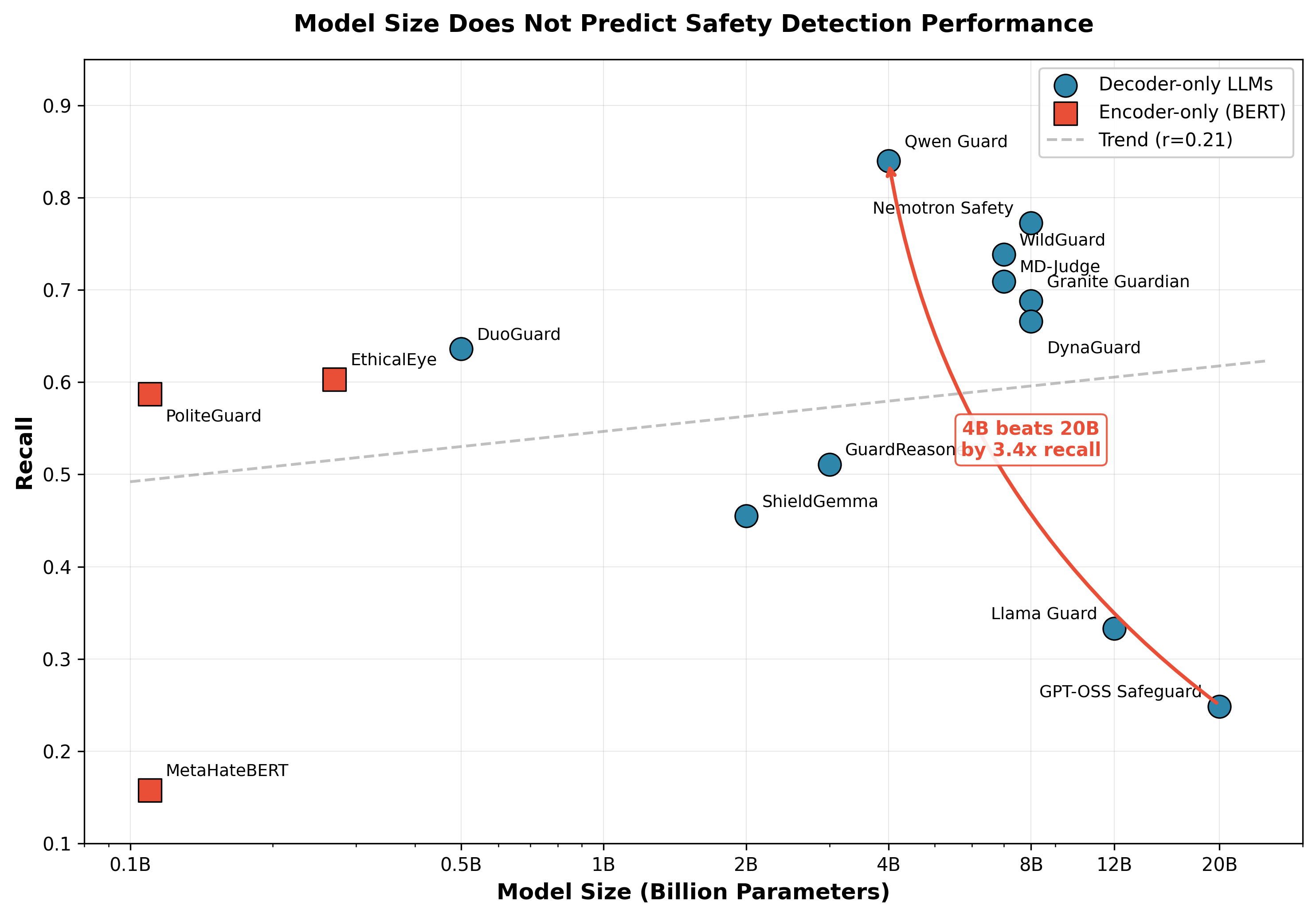}
\caption{Model size does not predict safety detection performance. Each point represents a guard model, with decoder-only LLMs (blue circles) and encoder-only transformers (red squares). The 4B Qwen Guard achieves 3.4x higher recall than the 20B GPT-OSS Safeguard. The weak trend line (r=0.21) confirms no meaningful correlation between model size and recall.}
\label{fig:size_vs_recall}
\end{figure}

The inverse correlation between model size and safety detection performance challenges conventional assumptions. Figure~\ref{fig:size_vs_recall} visualizes this relationship: Qwen Guard (4B) achieves 3.4x higher recall than GPT-OSS Safeguard (20B), and the Pearson correlation between log$_{10}$-transformed model size and recall is negligible ($r=0.21$, $p=0.48$, $n=14$). We hypothesize that larger models may be trained with more conservative safety thresholds, prioritizing avoiding false positives over comprehensive detection.

\textbf{Finding 3: Label normalization significantly impacts results.}

Qwen Guard outputs three labels: \textit{safe}, \textit{unsafe}, and \textit{controversial}. The \textit{controversial} category contains contextually harmful content (e.g., implicit threats, borderline hate speech). Table~\ref{tab:controversial_impact} shows the impact of treating \textit{controversial} as unsafe versus safe.

\begin{table}[t]
\centering
\small
\begin{tabular}{lccc}
\toprule
\textbf{Metric} & \textbf{Controv.=Safe} & \textbf{Controv.=Unsafe} & \textbf{$\Delta$} \\
\midrule
Recall & 46.75\% & 83.97\% & \textbf{+37.22\%} \\
Precision & 89.12\% & 68.79\% & --20.33\% \\
F1 Score & 61.33\% & 75.63\% & +14.30\% \\
\bottomrule
\end{tabular}
\caption{Impact of Qwen Guard's \textit{controversial} label treatment on evaluation metrics.}
\label{tab:controversial_impact}
\end{table}

Merging \textit{controversial} with \textit{unsafe} increases recall by 37.2 percentage points (from 46.75\% to 83.97\%) at the cost of 20.3 percentage points in precision. This tradeoff is justified for safety-critical applications: the F1 score improves by 14.3 percentage points, and-consistent with our central thesis-\textbf{catching 37\% more harmful content outweighs the increase in false positives}. We do not treat \textit{controversial} as a separate third class because our benchmark focuses on binary classification for production safety systems, where content must be either blocked or allowed-a three-class output does not provide actionable guidance for deployment. Without this merging, Qwen Guard's recall would drop to 46.75\%, ranking it 10th rather than 1st among evaluated models.

\subsection{Limitations}

Our study has limitations: (1) all safe samples originate from RealToxicityPrompts, potentially biasing evaluation; (2) the 0.5 threshold is a design choice, though model rankings remain stable across 0.3--0.7 (Section 4.4); (3) we evaluate only prompts, not responses; (4) evaluation is English-only; and (5) domain-specific applications may require specialized benchmarks.

\section{Conclusion}
\label{sec:conclusion}

We present the first comprehensive benchmark of 14 open-source safety guard models on 79,331 samples spanning 8 NIST safety categories, revealing that many widely-deployed models exhibit dangerous conservatism, missing up to 75\% of harmful content. Our key contributions are: (1) a methodology for NIST-aligned safety benchmarks; (2) empirical evidence that model size does not predict performance; and (3) actionable recommendations for model selection. Qwen Guard (4B) emerges as best-performing with 83.97\% recall, followed by Nemotron Safety (8B) and WildGuard (7B). Future work should address multilingual evaluation and response-level classification.

\subsubsection*{Reproducibility Statement}
All datasets are publicly available: HarmBench, StrongREJECT \citep{souly2024strongreject}, RealToxicityPrompts, and BeaverTails. Model inference used default configurations (temperature 0.0). Dataset construction follows Section~\ref{sec:dataset}.

\subsubsection*{Ethics Statement}
This work uses publicly available datasets containing harmful content for safety research. Our benchmark aims to improve AI safety. While metrics could inform attacks, transparent evaluation benefits outweigh risks.

\bibliography{iclr2026_conference}
\bibliographystyle{plainnat}

\appendix
\section{Additional Results}
\label{app:additional_results}

This appendix provides comprehensive supplementary results that support the main findings presented in the paper.

\subsection{Full Performance Metrics}

Table~\ref{tab:full_metrics} provides complete evaluation metrics for all 14 models, including ROC-AUC and Matthews Correlation Coefficient (MCC). MCC is particularly informative for imbalanced datasets as it considers all four confusion matrix quadrants. Models are ranked by recall, consistent with our thesis that detecting unsafe content is the primary objective for safety guard models. The MCC values reveal that even top-performing models show moderate correlation (0.40--0.46), indicating room for improvement across all evaluated systems.

\begin{table}[t]
\caption{Complete Evaluation Metrics}
\label{tab:full_metrics}
\begin{center}
\small
\begin{tabular}{lcccccc}
\toprule
\textbf{Model} & \textbf{Recall} & \textbf{Precision} & \textbf{F1} & \textbf{Acc} & \textbf{ROC-AUC} & \textbf{MCC} \\
\midrule
Qwen Guard & 0.8397 & 0.6879 & 0.7563 & 0.7036 & 0.6894 & 0.4004 \\
Nemotron Safety & 0.7725 & 0.7493 & 0.7607 & 0.7342 & 0.7302 & 0.4621 \\
WildGuard & 0.7383 & 0.7289 & 0.7336 & 0.7067 & 0.7034 & 0.4073 \\
MD-Judge & 0.7091 & 0.7316 & 0.7202 & 0.6986 & 0.6975 & 0.3940 \\
Granite Guardian & 0.6881 & 0.7677 & 0.7257 & 0.7155 & 0.7184 & 0.4349 \\
DynaGuard & 0.6659 & 0.7715 & 0.7148 & 0.7095 & 0.7140 & 0.4267 \\
DuoGuard & 0.6360 & 0.7612 & 0.6930 & 0.6918 & 0.6976 & 0.3949 \\
EthicalEye & 0.6032 & 0.6705 & 0.6350 & 0.6208 & 0.6226 & 0.2442 \\
PoliteGuard & 0.5873 & 0.5689 & 0.5780 & 0.5308 & 0.5250 & 0.0501 \\
GuardReasoner & 0.5109 & 0.7397 & 0.6044 & 0.6283 & 0.6430 & 0.2923 \\
ShieldGemma & 0.4549 & 0.8220 & 0.5857 & 0.6479 & 0.6680 & 0.3640 \\
Llama Guard & 0.3332 & 0.7851 & 0.4679 & 0.5860 & 0.6117 & 0.2637 \\
GPT-OSS Safeguard & 0.2486 & 0.8068 & 0.3800 & 0.5564 & 0.5883 & 0.2350 \\
MetaHateBERT & 0.1579 & 0.6541 & 0.2544 & 0.4937 & 0.5285 & 0.0839 \\
\bottomrule
\end{tabular}
\end{center}
\end{table}

\section{Implementation Details}
\label{app:implementation}

\subsection{Label Normalization Details}

\textbf{Qwen Guard} (Qwen3Guard-Gen-4B): Outputs three labels (safe, unsafe, controversial). The \textit{controversial} label is merged with \textit{unsafe} as these samples contain contextually harmful content.

\textbf{Nemotron Safety} (Llama-3.1-Nemotron-Safety-Guard-8B-v3): Returns JSON with ``User Safety'' and ``Safety Categories'' fields. We use the ``User Safety'' field for binary classification.

\textbf{WildGuard}: Returns structured output with ``Harmful request: yes/no''. We map ``yes'' to unsafe.

\textbf{MD-Judge} (MD-Judge-v0\_2-internlm2\_7b): LLM-based judge that outputs safe/unsafe classification with reasoning.

\textbf{Granite Guardian} (granite-guardian-3.3-8b): Returns safe/unsafe with optional violation categories.

\textbf{DynaGuard} (DynaGuard-4B): Decoder-based safety model returning binary safe/unsafe labels.

\textbf{DuoGuard} (DuoGuard-0.5B): Sequence classifier with 12 risk categories. Uses sigmoid activation; sample is unsafe if any category probability exceeds 0.5 threshold.

\textbf{EthicalEye}: Encoder-based classifier outputs Safe/Un-Safe. We map ``Un-Safe'' to unsafe.

\textbf{PoliteGuard}: Politeness classifier with four labels. We map ``impolite'' to unsafe; ``polite'', ``somewhat polite'', and ``neutral'' to safe.

\textbf{GuardReasoner} (GuardReasoner-3B): Llama-3.2-3B fine-tuned via R-SFT and HS-DPO. Uses simplified prompt for one-word safe/unsafe output.

\textbf{ShieldGemma} (shieldgemma-2b): Uses Yes/No generation for policy violation detection. ``Yes'' indicates policy violation (unsafe).

\textbf{Llama Guard} (Llama-Guard-4-12B): Returns safe/unsafe with S1--S14 category codes indicating violation type.

\textbf{GPT-OSS Safeguard}: Uses NIST category-specific policy prompts. Returns JSON with label and rationale.

\textbf{MetaHateBERT}: Hate speech encoder classifier. LABEL\_1 (hate) maps to unsafe; LABEL\_0 to safe.

\clearpage
\section{Dataset Statistics}
\label{app:dataset_stats}

\subsection{NIST Category Distribution}

\begin{table}[t]
\caption{Sample Distribution by NIST Safety Category}
\label{tab:nist_distribution}
\begin{center}
\begin{tabular}{lrr}
\toprule
\textbf{NIST Category} & \textbf{Count} & \textbf{Percentage} \\
\midrule
Harassment & 23,345 & 29.4\% \\
Threats & 17,187 & 21.7\% \\
Hate Speech & 13,697 & 17.3\% \\
Sexual Content & 11,084 & 14.0\% \\
Violence & 6,703 & 8.4\% \\
Profanity & 5,600 & 7.1\% \\
Health & 591 & 0.7\% \\
Suicide \& Self-Harm & 173 & 0.2\% \\
\midrule
\textbf{Total (Unsafe)} & \textbf{43,393} & \textbf{54.7\%} \\
Safe & 35,938 & 45.3\% \\
\midrule
\textbf{Grand Total} & \textbf{79,331} & \textbf{100\%} \\
\bottomrule
\end{tabular}
\end{center}
\end{table}

\noindent\textbf{Data Quality Notes.} BeaverTails is the only source for Suicide \& Self-Harm category (173 samples), while RealToxicityPrompts is the primary source for Profanity (5,600) and Threats (17,187). Safe samples exclusively come from RealToxicityPrompts due to other datasets being adversarial by design. Confidence scores are 1.0 for HarmBench, StrongREJECT, and BeaverTails, and variable (0.0--1.0) for RealToxicityPrompts.

\subsection{Threshold Sensitivity Analysis}
\label{app:threshold_sensitivity}

RealToxicityPrompts samples are labeled based on Perspective API toxicity scores. For each sample, we compute $\text{highest\_score} = \max(\text{combined scores across 7 dimensions})$, where combined score is the average of prompt and continuation scores per dimension. Samples are labeled unsafe if $\text{highest\_score} > \tau$. We analyze sensitivity to threshold $\tau \in \{0.3, 0.4, 0.5, 0.6, 0.7\}$.

\begin{table}[t]
\caption{Dataset Composition at Different Thresholds}
\label{tab:threshold_composition}
\begin{center}
\small
\begin{tabular}{lccccc}
\toprule
\textbf{Threshold} & \textbf{0.3} & \textbf{0.4} & \textbf{0.5} & \textbf{0.6} & \textbf{0.7} \\
\midrule
RTP Unsafe & 60,042 & 51,102 & 31,583 & 13,589 & 7,562 \\
RTP Safe & 7,479 & 16,419 & 35,938 & 53,932 & 59,959 \\
Other (always unsafe) & 11,810 & 11,810 & 11,810 & 11,810 & 11,810 \\
\midrule
\textbf{Total Unsafe} & 71,852 & 62,912 & 43,393 & 25,399 & 19,372 \\
\textbf{Unsafe \%} & 90.6\% & 79.3\% & 54.7\% & 32.0\% & 24.4\% \\
\bottomrule
\end{tabular}
\end{center}
\end{table}

\begin{table}[t]
\caption{Threshold Sensitivity: Recall, Precision, and F1 at Different Labeling Thresholds (All Models). Bold indicates threshold 0.5 (default). As threshold increases, ground truth becomes stricter-fewer samples are labeled unsafe-causing recall to increase and precision to decrease.}
\label{tab:threshold_sensitivity}
\begin{center}
\scriptsize
\begin{tabular}{llccccc}
\toprule
\textbf{Model} & \textbf{Metric} & \textbf{0.3} & \textbf{0.4} & \textbf{0.5} & \textbf{0.6} & \textbf{0.7} \\
\midrule
\multirow{3}{*}{Qwen Guard}
    & Recall & 0.712 & 0.757 & \textbf{0.840} & 0.898 & 0.923 \\
    & Precision & 0.965 & 0.899 & \textbf{0.688} & 0.431 & 0.338 \\
    & F1 & 0.819 & 0.822 & \textbf{0.756} & 0.582 & 0.494 \\
\midrule
\multirow{3}{*}{Nemotron Safety}
    & Recall & 0.613 & 0.675 & \textbf{0.773} & 0.836 & 0.868 \\
    & Precision & 0.985 & 0.949 & \textbf{0.749} & 0.474 & 0.376 \\
    & F1 & 0.756 & 0.789 & \textbf{0.761} & 0.605 & 0.525 \\
\midrule
\multirow{3}{*}{WildGuard}
    & Recall & 0.596 & 0.644 & \textbf{0.738} & 0.827 & 0.868 \\
    & Precision & 0.974 & 0.921 & \textbf{0.729} & 0.478 & 0.383 \\
    & F1 & 0.739 & 0.758 & \textbf{0.734} & 0.606 & 0.531 \\
\midrule
\multirow{3}{*}{MD-Judge}
    & Recall & 0.567 & 0.609 & \textbf{0.709} & 0.816 & 0.860 \\
    & Precision & 0.968 & 0.910 & \textbf{0.732} & 0.493 & 0.396 \\
    & F1 & 0.715 & 0.729 & \textbf{0.720} & 0.614 & 0.543 \\
\midrule
\multirow{3}{*}{Granite Guardian}
    & Recall & 0.533 & 0.588 & \textbf{0.688} & 0.779 & 0.832 \\
    & Precision & 0.985 & 0.952 & \textbf{0.768} & 0.509 & 0.414 \\
    & F1 & 0.692 & 0.727 & \textbf{0.726} & 0.616 & 0.553 \\
\midrule
\multirow{3}{*}{DynaGuard}
    & Recall & 0.511 & 0.559 & \textbf{0.666} & 0.764 & 0.805 \\
    & Precision & 0.980 & 0.939 & \textbf{0.772} & 0.518 & 0.417 \\
    & F1 & 0.672 & 0.701 & \textbf{0.715} & 0.618 & 0.550 \\
\midrule
\multirow{3}{*}{DuoGuard}
    & Recall & 0.498 & 0.552 & \textbf{0.636} & 0.715 & 0.767 \\
    & Precision & 0.986 & 0.957 & \textbf{0.761} & 0.501 & 0.410 \\
    & F1 & 0.662 & 0.700 & \textbf{0.693} & 0.589 & 0.535 \\
\midrule
\multirow{3}{*}{EthicalEye}
    & Recall & 0.533 & 0.578 & \textbf{0.603} & 0.543 & 0.496 \\
    & Precision & 0.982 & 0.931 & \textbf{0.670} & 0.353 & 0.246 \\
    & F1 & 0.691 & 0.713 & \textbf{0.635} & 0.428 & 0.329 \\
\midrule
\multirow{3}{*}{PoliteGuard}
    & Recall & 0.581 & 0.590 & \textbf{0.587} & 0.576 & 0.556 \\
    & Precision & 0.931 & 0.829 & \textbf{0.569} & 0.327 & 0.240 \\
    & F1 & 0.715 & 0.690 & \textbf{0.578} & 0.417 & 0.336 \\
\midrule
\multirow{3}{*}{GuardReasoner}
    & Recall & 0.406 & 0.432 & \textbf{0.511} & 0.622 & 0.686 \\
    & Precision & 0.964 & 0.903 & \textbf{0.740} & 0.525 & 0.439 \\
    & F1 & 0.571 & 0.584 & \textbf{0.604} & 0.569 & 0.535 \\
\midrule
\multirow{3}{*}{ShieldGemma}
    & Recall & 0.330 & 0.365 & \textbf{0.455} & 0.550 & 0.610 \\
    & Precision & 0.988 & 0.955 & \textbf{0.822} & 0.582 & 0.492 \\
    & F1 & 0.495 & 0.528 & \textbf{0.586} & 0.565 & 0.545 \\
\midrule
\multirow{3}{*}{Llama Guard}
    & Recall & 0.249 & 0.271 & \textbf{0.333} & 0.436 & 0.500 \\
    & Precision & 0.972 & 0.926 & \textbf{0.785} & 0.601 & 0.526 \\
    & F1 & 0.396 & 0.419 & \textbf{0.468} & 0.506 & 0.513 \\
\midrule
\multirow{3}{*}{GPT-OSS Safeguard}
    & Recall & 0.183 & 0.203 & \textbf{0.249} & 0.302 & 0.334 \\
    & Precision & 0.985 & 0.956 & \textbf{0.807} & 0.573 & 0.484 \\
    & F1 & 0.309 & 0.335 & \textbf{0.380} & 0.395 & 0.396 \\
\midrule
\multirow{3}{*}{MetaHateBERT}
    & Recall & 0.142 & 0.154 & \textbf{0.158} & 0.153 & 0.145 \\
    & Precision & 0.974 & 0.923 & \textbf{0.654} & 0.370 & 0.268 \\
    & F1 & 0.248 & 0.263 & \textbf{0.254} & 0.216 & 0.188 \\
\bottomrule
\end{tabular}
\end{center}
\end{table}

Key findings:
\begin{itemize}
    \item Ranking stability: Model rankings remain consistent across all thresholds. Qwen Guard maintains the highest recall regardless of threshold choice, followed by Nemotron Safety and WildGuard.
    \item Precision-recall tradeoff: As the threshold increases (stricter labeling), recall improves while precision degrades. At threshold 0.5, F1 scores peak for most models, indicating optimal balance.
    \item Anomalous behavior: EthicalEye, PoliteGuard, and MetaHateBERT show minimal recall variation across thresholds, suggesting these models' predictions are largely independent of subtle toxicity distinctions.
    \item Robustness: Our conclusions about relative model performance are robust to threshold selection.
\end{itemize}

\section{False Negative Analysis (Full Results)}
\label{app:fn_analysis}

This section provides the complete false negative analysis for all 14 models across all 8 NIST safety categories. False Negative Rate (FN\%) represents the percentage of unsafe samples incorrectly classified as safe-lower values indicate better detection. This comprehensive breakdown reveals category-specific strengths and weaknesses that are masked by aggregate metrics. For example, MD-Judge achieves the lowest FN rates on Violence (1.1\%) and Suicide (0.6\%), while EthicalEye excels on Harassment (10.5\%) and Profanity (7.7\%)-demonstrating that no single model dominates across all harm types.

\begin{table}[t]
\caption{False Negative Rate (\%) by NIST Category-All 14 Models. Models sorted by Overall FN rate (best to worst). Bold indicates best performer per category.}
\label{tab:fn_full}
\begin{center}
\scriptsize
\begin{tabular}{lcccccccc|c}
\toprule
\textbf{Model} & \textbf{Viol.} & \textbf{Suic.} & \textbf{Hate} & \textbf{Sex.} & \textbf{Health} & \textbf{Harass.} & \textbf{Prof.} & \textbf{Threat} & \textbf{Overall} \\
\midrule
Qwen Guard & 1.9 & 2.3 & \textbf{10.3} & \textbf{10.6} & 17.8 & 22.2 & 23.8 & \textbf{27.4} & \textbf{15.9} \\
Nemotron Safety & 4.4 & 6.9 & 29.3 & 17.0 & 26.2 & 24.2 & 15.1 & 38.8 & 22.8 \\
WildGuard & 3.1 & 4.0 & 18.9 & 27.1 & 22.2 & 29.6 & 44.1 & 42.3 & 26.2 \\
MD-Judge & \textbf{1.1} & \textbf{0.6} & 19.3 & 33.5 & \textbf{14.4} & 42.8 & 56.9 & 32.4 & 29.1 \\
Granite Guardian & 9.9 & 6.9 & 24.4 & 23.0 & 31.6 & 46.2 & 17.0 & 48.4 & 31.2 \\
DynaGuard & 8.7 & 5.2 & 23.9 & 31.6 & 26.9 & 45.4 & 51.7 & 43.4 & 33.2 \\
DuoGuard & 16.0 & 11.6 & 35.8 & 36.1 & 32.1 & 23.9 & 23.4 & 76.2 & 36.4 \\
GuardReasoner & 9.7 & 6.4 & 30.2 & 40.6 & 20.1 & 57.6 & 67.3 & 39.0 & 39.0 \\
EthicalEye & 81.4 & 58.4 & 51.7 & 22.5 & 78.5 & \textbf{10.5} & \textbf{7.7} & 55.8 & 39.7 \\
PoliteGuard & 59.9 & 46.8 & 30.4 & 59.7 & 65.7 & 16.0 & 17.4 & 55.8 & 41.3 \\
ShieldGemma & 30.3 & 4.0 & 35.0 & 35.7 & 45.2 & 83.5 & 85.6 & 65.7 & 54.5 \\
Llama Guard & 19.3 & 18.5 & 67.2 & 70.3 & 42.6 & 81.4 & 87.9 & 79.5 & 66.7 \\
GPT-OSS Safeguard & 63.5 & 38.7 & 65.9 & 73.2 & 78.0 & 77.1 & 91.4 & 88.6 & 75.1 \\
MetaHateBERT & 96.4 & 95.4 & 79.8 & 88.8 & 94.4 & 71.0 & 79.4 & 91.5 & 84.2 \\
\bottomrule
\end{tabular}
\end{center}
\end{table}

Extended observations:
\begin{itemize}
    \item Category difficulty: Violence and Suicide \& Self-Harm are consistently the easiest categories for detection (lowest FN rates), while Threats and Harassment show the highest variability across models.

    \item Specialized trade-offs: Some models excel in specific categories despite poor overall performance. EthicalEye achieves 7.7\% FN on Profanity (best among all models) but 81.4\% on Violence. PoliteGuard achieves 16.0\% on Harassment but 65.7\% on Health.

    \item Performance gap: The gap between best (Qwen Guard, 15.9\%) and worst (MetaHateBERT, 84.2\%) performers represents a 5.3$\times$ difference in false negative rate.

    \item Implicit vs. explicit harm: Models trained on explicit safety taxonomies (Qwen, Nemotron) consistently outperform those trained on general toxicity (MetaHateBERT, GPT-OSS) for contextually harmful content like Threats and Harassment.
\end{itemize}

\section{Per-Source Performance Analysis}
\label{app:per_source}

To verify that model rankings are not driven by dataset source artifacts, we analyze performance stratified by data source. RealToxicityPrompts (RTP) contains both safe and unsafe samples (67,521 samples), while adversarial datasets are 100\% unsafe by design. Tables~\ref{tab:f1_by_dataset}--\ref{tab:precision_by_dataset} show F1, Recall, and Precision by dataset for all 14 models.

\begin{table}[t]
\caption{F1 Score by Dataset (All 14 Models). Models sorted by overall recall.}
\label{tab:f1_by_dataset}
\begin{center}
\scriptsize
\begin{tabular}{lcccc}
\toprule
\textbf{Model} & \textbf{RTP} & \textbf{HarmBench} & \textbf{StrongREJECT} & \textbf{BeaverTails} \\
\midrule
Qwen Guard & 0.6892 & \textbf{1.0000} & 0.7059 & \textbf{0.9732} \\
Nemotron Safety & \textbf{0.7004} & 0.9951 & \textbf{1.0000} & 0.9378 \\
WildGuard & 0.6516 & \textbf{1.0000} & \textbf{1.0000} & 0.9644 \\
MD-Judge & 0.6281 & 0.9951 & \textbf{1.0000} & 0.9704 \\
Granite Guardian & 0.6536 & 0.9951 & \textbf{1.0000} & 0.9223 \\
DynaGuard & 0.6369 & 0.9902 & \textbf{1.0000} & 0.9208 \\
DuoGuard & 0.6127 & 0.8161 & 0.9000 & 0.9092 \\
EthicalEye & 0.6736 & 0.1273 & 0.3789 & 0.4711 \\
PoliteGuard & 0.5512 & 0.2833 & 0.4600 & 0.6881 \\
GuardReasoner & 0.4838 & 0.9949 & 0.9930 & 0.8995 \\
ShieldGemma & 0.5006 & 0.8984 & 0.9416 & 0.7901 \\
Llama Guard & 0.3269 & 0.9596 & 0.9630 & 0.7738 \\
GPT-OSS Safeguard & 0.3146 & 0.8925 & 0.9078 & 0.5330 \\
MetaHateBERT & 0.2770 & 0.0381 & 0.1446 & 0.1849 \\
\bottomrule
\end{tabular}
\end{center}
\end{table}
\begin{table}[t]
\caption{Recall by Dataset (All 14 Models). Models sorted by overall recall.}
\label{tab:recall_by_dataset}
\begin{center}
\scriptsize
\begin{tabular}{lcccc}
\toprule
\textbf{Model} & \textbf{RTP} & \textbf{HarmBench} & \textbf{StrongREJECT} & \textbf{BeaverTails} \\
\midrule
Qwen Guard & \textbf{0.8011} & \textbf{1.0000} & 0.5455 & \textbf{0.9477} \\
Nemotron Safety & 0.7303 & 0.9903 & \textbf{1.0000} & 0.8828 \\
WildGuard & 0.6656 & \textbf{1.0000} & \textbf{1.0000} & 0.9313 \\
MD-Judge & 0.6215 & 0.9903 & \textbf{1.0000} & 0.9424 \\
Granite Guardian & 0.6243 & 0.9903 & \textbf{1.0000} & 0.8559 \\
DynaGuard & 0.5942 & 0.9806 & \textbf{1.0000} & 0.8532 \\
DuoGuard & 0.5627 & 0.6893 & 0.8182 & 0.8335 \\
EthicalEye & 0.7146 & 0.0680 & 0.2338 & 0.3081 \\
PoliteGuard & 0.6131 & 0.1650 & 0.2987 & 0.5245 \\
GuardReasoner & 0.3974 & 0.9898 & 0.9860 & 0.8174 \\
ShieldGemma & 0.3791 & 0.8155 & 0.8896 & 0.6530 \\
Llama Guard & 0.2198 & 0.9223 & 0.9286 & 0.6310 \\
GPT-OSS Safeguard & 0.2019 & 0.8058 & 0.8312 & 0.3634 \\
MetaHateBERT & 0.1792 & 0.0194 & 0.0779 & 0.1019 \\
\bottomrule
\end{tabular}
\end{center}
\end{table}
\begin{table}[t]
\caption{Precision by Dataset (All 14 Models). Precision=1.0 for adversarial datasets because all samples are unsafe.}
\label{tab:precision_by_dataset}
\begin{center}
\scriptsize
\begin{tabular}{lcccc}
\toprule
\textbf{Model} & \textbf{RTP} & \textbf{HarmBench} & \textbf{StrongREJECT} & \textbf{BeaverTails} \\
\midrule
Qwen Guard & 0.6048 & 1.0 & 1.0 & 1.0 \\
Nemotron Safety & 0.6728 & 1.0 & 1.0 & 1.0 \\
WildGuard & 0.6382 & 1.0 & 1.0 & 1.0 \\
MD-Judge & 0.6348 & 1.0 & 1.0 & 1.0 \\
Granite Guardian & 0.6858 & 1.0 & 1.0 & 1.0 \\
DynaGuard & 0.6863 & 1.0 & 1.0 & 1.0 \\
DuoGuard & 0.6725 & 1.0 & 1.0 & 1.0 \\
EthicalEye & 0.6370 & 1.0 & 1.0 & 1.0 \\
PoliteGuard & 0.5006 & 1.0 & 1.0 & 1.0 \\
GuardReasoner & 0.6182 & 1.0 & 1.0 & 1.0 \\
ShieldGemma & \textbf{0.7369} & 1.0 & 1.0 & 1.0 \\
Llama Guard & 0.6375 & 1.0 & 1.0 & 1.0 \\
GPT-OSS Safeguard & 0.7117 & 1.0 & 1.0 & 1.0 \\
MetaHateBERT & 0.6097 & 1.0 & 1.0 & 1.0 \\
\bottomrule
\end{tabular}
\end{center}
\end{table}

\textbf{Key Findings:} (1) RTP is the most challenging source-all models show lower F1 on RTP vs adversarial datasets; (2) Model rankings are consistent across sources-Qwen Guard leads on RTP recall (80.1\%) and BeaverTails, while bottom models (MetaHateBERT, GPT-OSS) rank poorly everywhere; (3) Encoder models (EthicalEye, PoliteGuard) excel on RTP but fail on adversarial; (4) \textbf{Top-7 decoder LLMs maintain consistent rankings} across sources, confirming robustness to dataset composition.

\end{document}